%% file: dbcnn_iccv.tex
\documentclass[10pt,twocolumn,letterpaper]{article}

\usepackage{iccv}
\usepackage{times}
\usepackage{epsfig}
\usepackage{graphicx}
\usepackage{amsmath}
\usepackage{amssymb}
\usepackage[vlined,boxed,commentsnumbered]{algorithm2e}
\usepackage{subfigure}

\usepackage[pagebackref=true,breaklinks=true,letterpaper=true,colorlinks,bookmarks=false]{hyperref}

\iccvfinalcopy 

\def\iccvPaperID{655} 

\input{definitions}

\ificcvfinal\fi
\begin{document}

\title{Action Representation Using Classifier Decision Boundaries}

\author{
Jue Wang${^{1,3}}$\quad Anoop Cherian${^{2,3}}$\quad Fatih Porikli${^{1,2,3}}$\quad Stephen Gould${^{1,2}}$ \\
 ${^1}$Data61/CSIRO,\quad ${^2}$Australian Centre for Robotic Vision \\
 ${^3}$The Australian National University, Canberra, Australia\\
{\tt\small firstname.lastname@anu.edu.au}
}
\maketitle
\input{abstract}
\input{intro}

\input{related_work}

\input{proposed_method}

\input{expts}

\input{conclude}

{\small
\bibliographystyle{ieee}
\bibliography{dbcnn_cvpr}
}

\end{document}

%% file: definitions.tex
\newtheorem{definition}{Definition}

\newcommand{\reals}[1]{\mathbb{R}^{#1}}
\newcommand{\enorm}[1]{\left\|{#1}\right\|}

\newcommand{\set}[1]{\left\{#1\right\}}

\DeclareMathOperator*{\subjectto}{\text{subject to}}

\renewcommand\cdots{...}

\newcommand{\suptensor}[1]{\mathfrak{S}^{d}}

\DeclareMathOperator*{\argmin}{arg\,min}

\newcommand{\comment}[1]{}

\newcommand{\done}[1]{}
\newcommand{\actodo}[1]{}

\newcommand{\seq}{X}
\newcommand*{\pseq}[1]{X^{\small{+}}_{{#1}}}
\newcommand*{\hpseq}[1]{\hat{X}^{\small{+}}_{{#1}}}
\newcommand*{\nseq}[1]{X^{\small{-}}_{{#1}}}
\newcommand*{\pdataset}{\mathcal{X}^{\small{+}}_{}}
\newcommand*{\ndataset}{\mathcal{X}^{-}_{}}
\newcommand*{\nfeat}[2]{\mathbf{x}^{{#1}-}_{{#2}}}
\newcommand*{\pfeat}[2]{\mathbf{x}^{{#1}\small{+}}_{{#2}}}

\newcommand*{\feat}{\mathbf{x}}
\newcommand*{\hfeat}{\mathbf{\hat{x}}}
\newcommand*{\ypseq}[1]{y^{\small{+}}_{{#1}}}

\newcommand{\card}[1]{\left|{#1}\right|}
\DeclareMathOperator*{\svmp}{SVMP}
\DeclareMathOperator*{\nsvmp}{NSVMP}
\DeclareMathOperator*{\svm}{SVM}
\newcommand{\dataset}{\mathcal{X}}
\newcommand{\wbstack}{\left[\begin{array}{c}\!\!\!w\!\!\!\\\!\!\!b\!\!\!\end{array}\right]}

%% file: abstract.tex
\begin{abstract}
Most popular deep learning based models for action recognition are designed to generate separate predictions within their short temporal windows, which are often aggregated by heuristic means to assign an action label to the full video segment. Given that not all frames from a video characterize the underlying action, pooling schemes that impose equal importance to all frames might be unfavorable. 

In an attempt towards tackling this challenge, we propose a novel pooling scheme, dubbed~\emph{SVM pooling}, based on the notion that among the bag of features generated by a CNN on all temporal windows, there is at least one feature that characterizes the action. To this end, we learn a decision hyperplane that separates this unknown yet useful feature from the rest. Applying multiple instance learning in an SVM setup, we use the parameters of this separating hyperplane as a descriptor for the video. Since these parameters are directly related to the support vectors in a max-margin framework, they serve as robust representations for pooling of the CNN features. We devise a joint optimization objective and an efficient solver that learns these hyperplanes per video and the corresponding action classifiers over the hyperplanes. Showcased experiments on the standard HMDB and UCF101 datasets demonstrate state-of-the-art performance.

\end{abstract}

%% file: intro.tex
\section{Introduction}
\label{sec:intro}

We witness an astronomical increase of video data on the web. This data deluge has brought out the problem of effective representation of videos, specifically, their semantic content, to the forefront of computer vision research. The resurgence of deep convolutional neural networks, coexisting with the availability of high-performance computing platforms, has demonstrated significant progress in the performance of several problems in computer vision, and is now pushing forward its frontlines in action recognition and video understanding. However, current solutions are still far from being practically useful, arguably due to the volumetric nature of this data modality and the complex nature of real-world human actions~\cite{Wang2016,feichtenhofer2016convolutional, simonyan2014two, simonyan2014very}. 
 
\begin{figure}

	\begin{center}
        \includegraphics[width=1\linewidth,trim={0cm 0cm 5cm 0cm},clip]{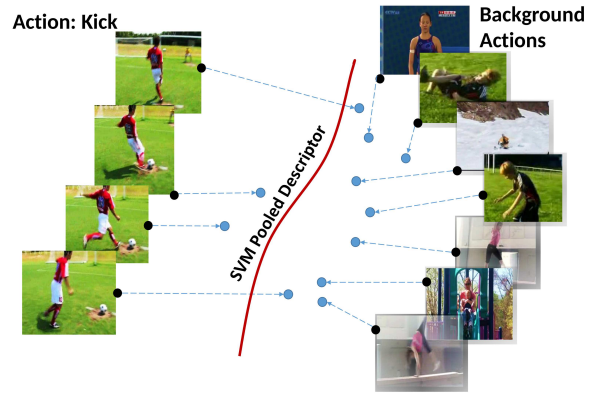}
	\end{center}
	\caption{Our scheme takes as input positive and negative bags of CNN features; the former capturing action sequences that we are interested in, and negative bags carrying background actions. We propose to learn a decision hyperplane in a multiple-instance learning framework that separates a predefined fraction of features in the positive bag against the negatives; the parameters of this decision boundary are used as the descriptor for the sequence, which is then used for action recognition.}
	\label{fig:1}
\end{figure}

Building on the availability of effective architectures, deep convolutional neural networks (CNNs) are often found to extract features from images that performs well on recognition tasks. Leveraging upon these performance benefits, deep solutions for video action recognition have been so far extensions of such image-based architectures. Nevertheless, video data could be of arbitrary length and scaling up image based CNN architectures to yet another dimension of complexity is not an easy task as the number of parameters in such models will be significantly higher. This demands greater computational infrastructures and large quantities of clean training data. Instead, the trend has been on converting the video data to short temporal segments consisting of one to a few frames, on which the available image-based CNN models could be trained. For example, in the popular two-stream CNN model for action recognition~\cite{feichtenhofer2016convolutional, wangtwo, simonyan2014two, simonyan2014very, wang2015action}, video data is split in two independent streams, one taking single RGB video frames, and the other using a small stack of optical flow images. These streams produce action predictions from such short video snippets, later the results are pooled to generate a prediction for the full sequence. 

Typically, the CNN classifier scores from video subsequences are combined using average pooling, max pooling, or are fused using a linear SVM. While average pooling gives equal weights to all the scores, max pooling may be sensitive to outlier predictions, and SVM may be confounded by predictions from background actions. Several works try to tackle this problem by using different pooling strategies \cite{bilen2016dynamic, fernando2015modeling, wang2015action,yue2015beyond,Wang2016}, which achieve some improvement compared with the baseline algorithm. 

To this end, we observe that not all CNN predictions are equally informative to recognize the activity, but at least some of the them are~\cite{schindler2008action}. This allows us to cast the problem in a multiple instance learning (MIL) framework, where we assume that some of the features from the intermediate layers of a CNN, trained to recognize actions, are indeed useful, while the rest of the features are not so. As shown in Figure \ref{fig:1}, we assume such a bag of features containing both the good and bad features to represent a positive class, while features from a known but different set of actions is used as a negative class. We then formulate a binary classification problem of separating as much good features as possible using an SVM. The classifier decision boundary learned is then used as a descriptor for the respective video sequence, dubbed the SVM Pooled (SVMP) descriptor. Consequently, this descriptor is used in an action classification setup. We also provide a joint objective that learns both the SVMP descriptors and the action classifiers. 

Against popular pooling schemes, our proposed framework offers several benefits, namely (i) it produces a compact representation of a video sequence of arbitrary length by characterizing the classifiability of its features against a selected background set, (ii) it is robust to classifier outliers as accounted for by the SVM formulation, and (iii) is computationally efficient \actodo{any other benefits}.

We provide extensive experimental evidence on two benchmark action recognition datasets, namely (i) the HMDB-51 dataset and (ii) the UCF101 dataset. Our results show that using SVMP descriptors improves the action recognition performance significantly. For example, it leads to about 6.3\% improvement on the HMDB-51\cite{kuehne2011hmdb} and 1.5\% on the UCF-101\cite{soomro2012ucf101} dataset in comparison to using original CNN features with popular pooling schemes. Further, when combined with other hand-crafted features, our results show state-of-the-art performance on the HMDB 51, while demonstrating promising results on UCF101. Given the simplicity of our approach against the challenging nature of these datasets, we believe the benefits afforded by our scheme is a significant step towards the advancement of action recognition.

Before reviewing related works in the next section, we summarize below the primary contributions of this paper:
\begin{itemize}
	\item We introduce an efficient and powerful pooling scheme, SVM pooling, for summarizing actions in video sequences that can filter useful features from a bag of per-frame CNN features. We contrive to use the output of SVM pooling as a descriptor for representing the sequence, which we call the SVM Pooled (SVMP) descriptor.
	
    \vspace{-2mm}
	\item We propose a joint optimization problem for learning SVMP descriptor per sequence and action classifiers on these descriptors, and propose an efficient block-coordinate descent scheme for solving it.

	\vspace{-2mm}
	\item We provide extensions of the SVMP descriptor to deal with non-linear sequence level decision boundaries. We also devise a multiple kernel scheme for fusing the linear and non-linear SVMP descriptors for better action recognition.
	
    \vspace{-2mm}
	\item Further, we report extensive experimental comparisons demonstrating the effectiveness of our scheme on two challenging benchmark datasets and demonstrate the state-of-the-art performance.
\end{itemize}

%% file: related_work.tex
\begin{figure*}[ht]
	\begin{center}
        \includegraphics[width=1\linewidth,trim={0cm 0cm 0cm 0cm},clip]{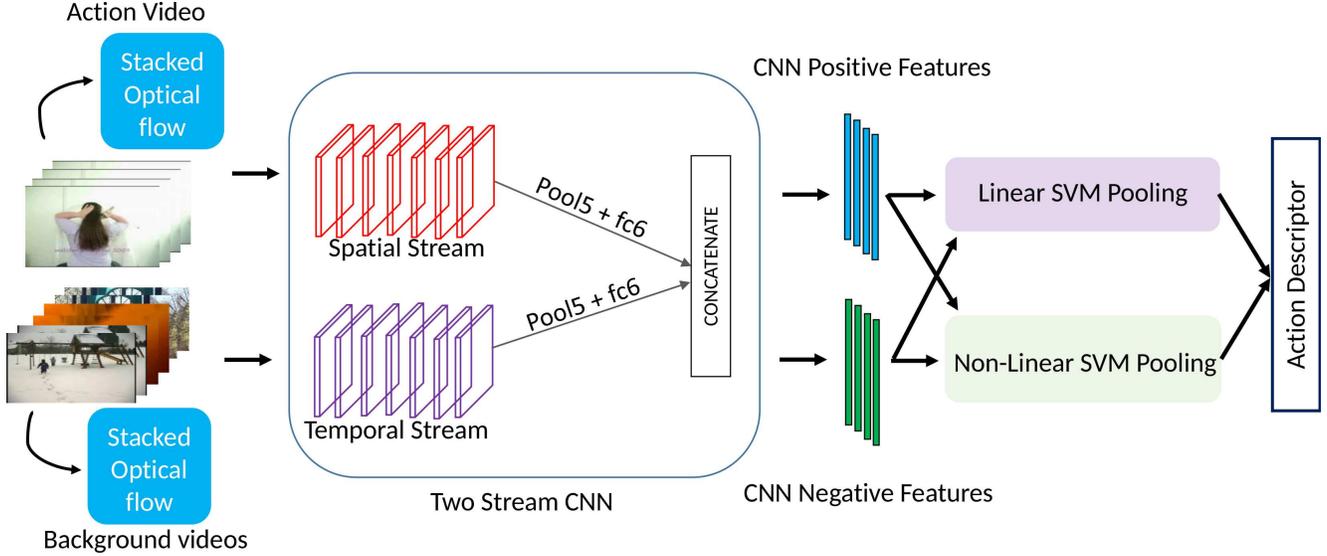}
	\end{center}
	\caption{Illustration of our SVM Pooling scheme. We take video sequences from action and background sequences, and aggregate the features generated by a two-stream CNN model using our pooling scheme into a single action descriptor.}
   	\label{fig:2}
\end{figure*}

\section{Related Work}
\label{sec:related_work}
Traditional methods for video action recognition typically use hand-crafted local features, such as dense trajectories, HOG, HOF, etc.~\cite{wang2013action}, or mid-level representations on them, such as Fisher Vectors~\cite{sadanand2012action}. With the resurgence of deep learning methods for object recognition~\cite{krizhevsky2012imagenet}, there have been several attempts to adapt these models to action recognition. One of the most successful is the two-stream CNN model proposed in ~\cite{simonyan2014two} which decouples the  spatial and temporal streams, thereby learning context and action dynamics separately. These streams are trained densely and independently; and at test time, their predictions are pooled. There have been extensions to this basic architecture using deeper networks and fusion of intermediate CNN layers~\cite{feichtenhofer2016convolutional,feichtenhofer2016spatiotemporal}. However, it is often argued that such dense sampling is inefficient and redundant; and only a few snippets are sufficient for recognition~\cite{schindler2008action}. Towards this end, we resort to a pooling scheme for selecting informative features generated by the two-stream model for generating our action descriptor; this selection is automatically decided by our MIL framework. A similar scheme is proposed in~\cite{Wang2016}, however they use manually-defined  video segmentation for equally-spaced snippet sampling.

Typically, pooling schemes consolidate input data into compact representations. Instead, we use the parameters of the data modeling function as the representation. 
This is similar to the recently proposed temporal pooling schemes such as rank pooling~\cite{fernando2015modeling,fernando2016discriminative}, dynamic images~\cite{bilen2016dynamic}, and dynamic flow~\cite{dynamic_flow}. However, while these methods optimize a rank-SVM based regression formulation, our motivation and formulation are different. We use the parameters of a binary SVM, which is trained to classify the video features from a pre-selected (but arbitrary) bag of negative features. In this respect, our pooling scheme is also different from Exemplar-SVMs~\cite{malisiewicz2011ensemble,willems2009exemplar,zepeda2015exemplar} that learns feature filters per data sample and then use these filters for feature extraction. While, it may seem that our scheme may lose the temporal aspect of actions, recall that these cues are already encoded in our two-stream generated features.

An important component of our scheme is the MIL scheme, which is a popular data selection technique~\cite{cinbis2017weakly,zhang2015self,wu2015deep,li2015multiple,yi2016human}. In the context of action recognition, schemes similar in motivation have been suggested. For example, Satkin and Hebert~\cite{satkin2010modeling} explores the effect of temporal cropping of videos to regions of actions; however assumes these regions are continuous. Nowozin et al.~\cite{nowozin2007discriminative} represents videos as sequences of discretized spatio-temporal sets and reduces the recognition problem into a max-gain sequence finding problem on these sets using an LPBoost classifier. Similar to ours, Li et al.~\cite{li2013dynamic} proposes an MIL setup for complex activity recognition using a dynamic pooling operator--a binary vector that selects input frames to be part of an action, which is learned by reducing the MIL problem to a set of linear programs. In~\cite{sun2014discover}, Chen and Nevatia propose a latent variable based model to explicitly localize discriminative video segments where events take place. In Vahdat et al.~\cite{vahdat2013compositional}, a compositional model for video event detection is presented using a multiple kernel learning based latent SVM. While all these schemes share similar motivations as ours, we cast our MIL problem in the setting of normalized set kernels~\cite{gartner2002multi} and reduce the formulation to standard SVM setup which can be solved rapidly. 

We note that there have been several other popular deep learning models devised for action modeling such as using 3D convolutional filters~\cite{tran2015learning}, recurrent neural networks~\cite{baccouche2011sequential,du2015hierarchical} and long-short term memory networks~\cite{donahue2015long,li2016action,yue2015beyond,srivastava2015unsupervised}. While, using recurrent architectures may benefit action recognition, datasets for this task are often small or are noisy, that training is usually difficult. 

%% file: proposed_method.tex
\section{Proposed Method}
\label{sec:proposed_method}
In this section, we describe our proposed method for learning SVMP descriptors and the action classifiers on them.~\actodo{Give a brief overview of the proposed method before going into details.} The overall pipeline is presented in the Figure \ref{fig:2}. 

\subsection{Problem Setup}
Let us assume we are given a dataset of $N$ video sequences $\pdataset = \set{\pseq{1}, \pseq{2},\cdots, \pseq{N}}$, where each $\pseq{i}$ is a set of frame level features, $\ie$, $\pseq{i}=\set{\pfeat{i}{1}, \pfeat{i}{2}, \cdots, \pfeat{i}{n}}$, each $\pfeat{i}{k}\in\reals{p}$. We assume that each $\pseq{i}$ is associated with an action class label $\ypseq{i}\in\set{1,2,\cdots, d}$. Further, the $+$ sign denotes that the features and the sequences represent a positive bag. We also assume that we have access to a set of sequences $\ndataset=\set{\nseq{1}, \nseq{2},\cdots \nseq{M}}$ belonging to actions different from those in $\pdataset$, where each $\nseq{j}=\set{\nfeat{j}{1}, \nfeat{j}{2}, \cdots, \nfeat{j}{n}}$ are the features associated with a negative bag, where each $\nfeat{j}{k}\in\reals{p}$. For simplicity, we assume all sequences have same number $n$ of features.

Our goals are two-fold, namely (i) to learn a classifier decision boundary for every sequence in $\pdataset$ that separates a fraction $\eta$ of them from the features in $\ndataset$ and (ii) to learn action classifiers on the classes in the positive bags that are represented by the learned decision boundaries in (i). In the following, we first provide a multiple instance learning formulation for achieving (i), and a joint objective combining (i) and learning (ii).

\subsection{Learning Decision Boundaries}
As described above, our goal in this section is to generate a descriptor for each sequence $\pseq{}\in\pdataset$; this descriptor we define to be the learned parameters of a hyperplane that separates the features $\pfeat{}{}\in\pseq{}$ from all features in $\ndataset$ . We do not want to warrant that all $\pfeat{}{}$ can be separated from $\ndataset$ (since several of them may belong to a background class), however we assume that at least a fixed fraction $\eta$ of them are classifiable. Mathematically, suppose the tuple $(w_i,b_i)$ represents the parameters of a max-margin hyperplane separating some of the features in a positive bag $\pseq{i}$ from all features in $\ndataset$, then we cast the following objective, which is a variant of the sparse MIL (SMIL)~\cite{bunescu2007multiple} and normalized set kernel (NSK)~\cite{gartner2002multi} formulations:
\begin{align}
\label{eq:mil}
\argmin_{w_i\in\reals{p},b_i\in\reals{},\zeta\geq 0} P1(w_i,b_i) := \enorm{w_i}^2 + & C_1\sum_{k=1}^{(M+1)n}\zeta_k\\
\subjectto \quad \theta(\feat;\eta)\left(w_i^T\feat+b_i\right) & \geq 1 - \zeta_k\\
\label{eq:5}\theta(\feat;\eta)= -1, \quad  \forall \feat \in \left\{\pseq{i}\bigcup \ndataset\right\}&\backslash \hpseq{i}\\
\label{eq:6}\theta(\hfeat; \eta) = 1, \quad \forall \hfeat \in\hpseq{i} & \\
\frac{\card{\hpseq{i}}}{\card{\pseq{i}}} \geq \eta. \quad\quad&
\label{eq:ratio-constraint} 
\end{align} 
In the above formulation, we assume that there is a subset $\hpseq{i}\subset\pseq{i}$ that is classifiable, while the rest of the positive bag need not be, as captured by the ratio in~\eqref{eq:ratio-constraint}. The variables $\zeta$ capture the non-negative slacks weighted by a regularization parameter $C_1$, and the function $\theta$ provides the label of the respective features. Unlike SMIL or NSK objectives, that assumes the individual features $\feat$ are summable, our problem is non-convex due to the unknown set $\hpseq{}$. However, this is not a serious deterrent to the usefulness of our formulation and can be tackled easily as described in the sequel and supported by our experimental results.

Given that the above formulation is built on an SVM objective, we call the scheme~\emph{SVM pooling} and formally define the descriptor for a sequence as follows:
\begin{definition}[SVM Pooling Descriptor]
\label{def:svmp}
Given a sequence $\seq$ of features $\feat\in\reals{p}$ and a negative dataset $\ndataset$, we define the~\emph{SVM Pooling} (SVMP) descriptor as $\svmp(\seq) = [w,b]^T\in\reals{p+1}$, where the tuple $(w,b)$ is obtained as the solution of problem $P1$ defined in~\eqref{eq:mil}.
\end{definition}

\subsection{Learning Action Classifiers}
Given a dataset of action sequences $\pdataset$ and a negative bag $\ndataset$, we propose to learn the SVMP descriptors per action sequence and the  classifiers on $\pdataset$ jointly as multi-class structured SVM problem which includes the MIL problem $P1$ as a sub-objective. The joint formulation is as follows:
\begin{equation}
\min_{w,b,Z} P2 :=\sum_{i=1}^N \enorm{w_i}^2\!\! +  \sum_{j=1}^d \enorm{Z_j}^2 
			 + C_2\sum_{i,l=1}^N \gamma_{il} + C_1\sum_{i=1}^{N}\hspace*{-0.05cm}\sum_{k=1}^{(M+1)n} \hspace*{-0.3cm}\zeta_{ik}\nonumber
\end{equation}
\vspace*{-0.5cm}
\begin{align}
\label{eq:9}
Z_j^T\left(\wbstack_i\!\!\!-\!\!\wbstack_l\right) & \geq \Delta(\ypseq{i},\ypseq{l}) - \gamma_{il},\ \ypseq{i}\!\!=\!j,  \forall\! \ypseq{l}\in\!\set{1,\cdots, d}
\end{align}
\vspace*{-0.5cm}
\begin{align}
\theta(\feat; \eta)\left(w_i^T\feat+b_i\right)  &\geq 1 - \zeta_{ik},\  \feat\in\pseq{i}\bigcup\ndataset, \forall i=1,\cdots, N\nonumber\\ 
\text{and}\quad \theta(\feat; \eta) &\in\set{+1,-1}, \gamma_{il}\geq 0,\ \xi_{ik}\geq 0,\nonumber
\end{align}
where $\theta(x.;\eta)$ is as defined in~\eqref{eq:5} and~\eqref{eq:6}. The function $\Delta(y,z)$ computes the similarity between the action labels $y$ and $z$. The above formulation P2 jointly optimizes the computations of SVMP descriptors per sequence and the parameters $Z$ of $d$ action classifiers, in a one-versus-rest fashion as described in~\eqref{eq:9}. The constant $C_2$ is a regularization parameter on the action classifiers and $\gamma$ represents the respective slack variables per sequence. 

\subsection{Efficient Optimization}
The problem $P2$ is not convex due to the function $\theta(\feat;\eta)$ that needs to select a set from the positive bags that satisfy the criteria in~\eqref{eq:ratio-constraint}. Also, note that the sub-problem $P1$ could be posed as a mixed-integer quadratic program (MIQP), which is known to be in NP~\cite{lazimy1982mixed}. While, there are efficient approximate solutions for this problem (such as~\cite{misener2013glomiqo}), the solution must be scalable to large number of high-dimensional features generated by a CNN. To this end, we propose the following relaxation.

Note that the regularization parameter $C_1$ in~\eqref{eq:mil} controls the positiveness of the slack variables $\zeta$, thereby influencing the training error rate. A smaller value of $C_1$ allows more data points to be misclassified. If we make the assumption that useful features from the sequences are easily classifiable compared to background features, then a smaller value of $C_1$ could help find the decision hyperplane easily. However, the correct value of $C_1$ depends on each sequence. Thus, in Algorithm~\eqref{alg1}, we propose a heuristic scheme to find the SVMP descriptor for a given sequence $\pseq{}$ by iteratively tuning $C_1$ such that at least a fraction $\eta$ of the features in the positive bag are classified as positive. 

\begin{algorithm}  
	\SetAlgoLined
	\KwIn{$\pseq{}$, $\ndataset$, $\eta$}
	$C_1 \leftarrow \epsilon,\ \lambda > 0$\;
	\Repeat{$\frac{\card{\hpseq{}}}{\card{\pseq{}}}\geq \eta$} {
		$C_1 \leftarrow \lambda C_1$\;
		$[w,b] \leftarrow \argmin_{w,b} \svm(\pseq{},\ \ndataset,\ C_1)$\;
		$\hpseq{} \leftarrow \set{\feat\in\pseq{}\ |\ w^T\feat+b \geq 0}$\;
	}
	\KwRet{$[w,b]$}
	\caption{Efficient solution to the MIL problem P1.}
	\label{alg1}
\end{algorithm}
 
Each step of Algorithm~\eqref{alg1} solves a standard SVM objective. Suppose we have an oracle that could give us a fixed value $C$ for $C_1$ that works for all action sequences for a fixed $\eta$. As is clear, there could be multiple combinations of data points in $\hpseq{}$ that could satisfy this $\eta$. If $\hpseq{p}$ is one such $\hpseq{}$. Then, P1 using $\hpseq{p}$ is just the SVM formulation and is thus convex. That is, if we enumerate all such $\hpseq{p}$ that satisfies the constraint using $\eta$, then the objective for each such $\hpseq{p}$ is an SVM problem, that could be solved using standard efficient solvers. Instead of enumerating all such bags $\hpseq{p}$, in Alg.~\ref{alg1}, we adjust the SVM classification rate to $\eta$, which is easier to implement. Assuming we find a $C_1$ that satisfies the $\eta$-constraint using P1, then due to the convexity of SVM, it can be shown that the optimizing objective of P1 will be the same in both cases (exhaustive enumeration and our proposed regularization adjustment), albeit the solution $\hat{X}_p^+$ might differ (there could be multiple solutions). 

Considering P2, it is non-convex in $Z$ and $(w_i,b_i)$'s jointly. However, it is convex in $Z$ when fixing $(w_i,b_i),\forall i\in\set{1,2,\cdots, N}$. Thus, under the above conditions, if we need to run only one iteration of P1, then P2 becomes convex in either variables separately, and thus we could solve it using block coordinate descent (BCD) towards a local minimum. Algorithm~\ref{alg2} depicts the iterations. Note that there is a coupling between the data point decision boundaries $(w_i,b_i)$ and the action classifier decision boundaries $Z_j$ in~\eqref{eq:9}, either of which are fixed when optimizing over the other using BCD. When optimizing over $(w_i,b_i)$, $Z_j^T\wbstack_l$ (in~\eqref{eq:9}) is a constant, and we use $\Delta(y_i^+, y_i^+)=1$, in which case the problem is equivalent to assuming $Z$ as a~\emph{virtual} positive data point in the positive bag. We make use of this observation in Algorithm~\ref{alg2} by including $Z$ in the positive bag. Note that these virtual $Z$ points are updated in place rather than adding new points in every iteration.


\begin{algorithm}  
	\SetAlgoLined
	\KwIn{$\pdataset$, $\ndataset$, $\eta$}
	\Repeat{until convergence} {
		\tcc{compute SVMP descriptors for all sequences}
		\For{$\pseq{i}\in\pdataset$}
		{
			$[w_i,b_i] \leftarrow \argmin_{w,b} \svm(\pseq{i},\ \ndataset,\ C)$\;
		}
		$Z\ \leftarrow\ \text{Solve P2 fixing $\wbstack_i, \forall i=\set{1,\cdots, N}$}$\;
		\tcc{$Z$ is added to $\pseq{i}$ so that $\svm$ could be used to satify~\eqref{eq:9}}
		$\pseq{i}\leftarrow \pseq{i} \cup Z$ 	
	}
	\KwRet{$Z$}
	\caption{A block-coordinate scheme for P2.}
	\label{alg2}
\end{algorithm}

When using decision boundaries as data descriptors, a natural question can be regarding the identifiability of the sequences using this descriptor, especially if the negative bag is randomly sampled. To circumvent this issue, we propose two workarounds, namely (i) to use the same negative bag for all the sequence, and (ii) assume all features (including positives and negatives) are centralized with respect to a global data mean.

\subsection{Kernelized Extensions}
In problem P1, we assume a linear decision boundary generating SVMP descriptors. However, looking back at our solutions in Algorithms~\eqref{alg1} and~\eqref{alg2}, it is clear that we deal with standard SVM formulations to solve our relaxed objectives. In the light of this, instead of using linear hyperplanes for classification, we may use non-linear decision boundaries by using the kernel trick to embed the data in a Hilbert space for better representation. Specifically, we propose RBF kernels for P1. Assuming $\dataset=\pdataset\cup\ndataset$, by the representation theorem~\cite{smola1998learning}, it is well-known that for a kernel $K:\dataset\times \dataset\rightarrow \reals{}_+$, the decision function $f$ for the SVM problem P1 will be of the form:

\begin{equation}
f(.) = \sum_{\feat\in\pseq{}\cup\ndataset}\alpha_{\feat} K(., \feat),
\label{eq:ksvm}
\end{equation}
where $\alpha_{\feat}$ are the parameters of the non-linear decision boundaries. Thus, instead of the linear SVMP descriptors, we could use the vector of $\alpha_{\feat}$ to describe the actions in a sequence. We call such a descriptor, a~\emph{non-linear SVM pooling} (NSVMP) descriptor. 

\subsection{Multi-Kernel Fusion}
We note that the problem P2 also reduces to a structured SVM formulation and thus we could substitute a non-linear kernel for action classifiers. Even better, we could learn classifiers over the non-linear NSVMP descriptors. Here we propose to use both SVMP and NSVMP descriptors via a multiple kernel fusion. That is, suppose $K_{\svmp}$ and $K_{\nsvmp}$ are two kernels constructed on the same data set of sequences $\pseq{}\in\pdataset$, then we propose to use a new kernel $K$:
\begin{equation}
K = \beta_1 K_{\svmp} + \beta_2 K_{\nsvmp},\quad \beta_1,\beta_2\geq 0.
\label{eq:mkl}
\end{equation}

Given that NSVMP descriptors are non-linear, we found an RBF works better for action classification, while a linear kernel seemed to provide good performance for SVMP descriptors. We also found that using homogeneous kernel linearization \cite{vedaldi2012efficient} on the NSVMP descriptors is useful. 

%% file: expts.tex
\section{Experiments}
\label{sec:exp}
We first introduce the datasets used in our experiments, followed by an exposition to the implementation details of our framework, analysis of the performance of each module, and extensive comparisons to previous works.

\subsection{Datasets}
We evaluate the performance of our scheme on two standard action recognition benchmarks, namely (i) HMDB-51~\cite{kuehne2011hmdb}, and the UCF101 datasets~\cite{soomro2012ucf101}. As the name implies, HMDB-51 dataset consists of 51 action classes and 6766 videos. The UCF101 dataset is double the size with 13320 videos and 101 actions. Both these datasets are evaluated using 3-fold cross-validation and mean classification accuracy is reported.



\subsection{CNN Model Training and Feature Extraction}
Due to well-known performance benefits, as well as, relative ease of training, we use a two-stream CNN architecture for action recognition. We use the VGG-16 and the ResNet-152 model for our data streams~\cite{simonyan2014very,feichtenhofer2016convolutional}. For the UCF101 dataset, we directly use publicly available models from ~\cite{feichtenhofer2016convolutional}. Note that, even though we use a VGG/ResNet model in our framework, our scheme is general and could use any other features or CNN architectures. For the HMDB dataset, we fine-tune a two-stream VGG model from the respective UCF101 model.

In the two-stream model, the spatial stream uses single RGB frames as input. To this end, we first resize the video frames to make the smaller side equal to 256. Further, we augment all frames via random horizontal flips and random crops to a 224 x 224 region. For temporal stream, which takes a stack of optical flow images as input, we choose the OpenCV implementation of the TV-L1 algorithm for flow computation~\cite{zach2007duality}. This follows generating a 10-channel stack of flow images as input to the CNN models. For every flow image stack, we subtract the median value, to reduce the impact of camera motion, followed by thresholding them in the range of $\pm$20 pixels and setting every other flow vector outside this range to zero, thereby removing outliers.

As alluded to above, for the HMDB dataset, both spatial and temporal networks are fine-tuned from the respective UCF101 models with an initial learning rate of $10^{-4}$ and high drop-out of 0.85 for the fully-connected 'fc6' and 0.9 for 'fc7' CNN layers (VGG model) as recommended in \cite{simonyan2014very,feichtenhofer2016convolutional}. To prevent over-fitting, we subsequently increase the drop-out once the validation loss begins to increase. The network is trained using SGD with a momentum of 0.9 and weight decay of 0.0005. We use mini-batches of size 64 for training. For computing the decision boundaries, we use the features from the intermediate CNN layers as explained in the sequel. 

\subsection{Selecting Negative Bags}
An important step in our algorithm is the selection of the positive and negative bags in the MIL problem. We randomly sample the required number (25) of frames from each sequence in the training/testing set to define the positive bags. For the negative bags, we need to select sequences that are unrelated to the actions in our evaluation datasets. We explored four different negatives in this regard to understand the impact of this selection, namely (when evaluating on HMDB-51 and vice versa on UCF101) using sequences from (i) ActivityNet dataset~\cite{caba2015activitynet}  unrelated to actions in HMDB-51, (ii) UCF101 (unrelated to HMDB-51), (iii) Thumos Challenge background sequences\footnote{http://www.thumos.info/home.html}, and (iv) synthesized random white noise sequences. For (i) and (ii), we use 50 frames each from randomly selected videos, one from every unrelated class, and for (iv) we used 50 synthesized white noise images, and randomly generated stack of optical flow images. As shown in Figure~\ref{fig:8} and~\ref{fig:3}, while there is a significant difference in the behaviour of our scheme when using the white noise bag, it appears to behave similarly for high values of $\eta$ parameter, implying that the our SVMP descriptor is indeed learning those dimensions of the CNN features that are informative for pooling and subsequent classification. Since ActivityNet negative bags showed the best performance, we use it in subsequent experiments.


\subsection{Results}
We first empirically analyze the benefits afforded by each component in our algorithm, namely (i) the choice of the intermediate layers used for creating our SVMP descriptor, (ii) choice of key parameters, (iii) benefits of SVMP over its non-linear variant NSVMP, (iv) benefit of using SVMP as against using standard pooling of CNN features, and (v) comparisons to the state of the art. 

\subsubsection{SVMP from Different CNN Features:} 
In this experiment, we generate SVMP descriptors from different intermediate layers of the CNN and compare their performance. In Table~\ref{table:1} and \ref{table:1.5}, we show this comparison on the split-1 of HMDB51 dataset and UCF101 dataset. We evaluate the results on pool5, fc6, fc7, fc8, and the softmax output layer using the VGG-16 model and pool5 and fc1000 in the ResNet-152 model. Specifically, features from each of these layers are used as the positive bags, and decision boundaries are computed using Algorithm~\ref{alg1} and~\ref{alg2} against the chosen set of negative bags. These decision boundaries are then used as the SVMP descriptors for the sequence and classified. The second column of Table~\ref{table:1} and \ref{table:1.5} shows the performance of the SVMP descriptor for each of the CNN features. As seen from the table, fc6 in VGG and pool5 in ResNet features show the best performance, better by about 6\% over the next best features. We further investigated for any complementary benefits that these features offer against other layers. In the third column in Table~\ref{table:1} and \ref{table:1.5}, we show the accuracy when combining features from other layers with fc6 and pool5. Interestingly, we find that fc6 combined with pool5 performs better in VGG and pool5 alone works better in ResNet. Thus, we use these combinations for the respective models in our experiments. 

\begin{table}[]
\centering
\caption{Comparison of SVMP descriptors on various CNN Features in VGG model over HMDB split 1.}
\label{table:1}
\begin{tabular}{lcc}
\hline
Layer   & Accuracy   & Accuracy\\
        & independently & combined with fc6\\
\hline
pool5   & 57.90\%         & \textbf{63.79\%}                                  \\
fc6     & \textbf{63.33\%}      & N.A                               \\
fc7     & 56.14\%         & 57.05\%                                  \\
fc8     & 52.35\%         & 58.62\%                                  \\
softmax & 41.04\%         & 46.23\%                                 
\end{tabular}
\end{table}

\begin{table}[]
\centering
\caption{Comparison of SVMP descriptors on various CNN Features in ResNet model over UCF101 split 1.}
\label{table:1.5}
\begin{tabular}{lcc}
\hline
Layer   & Accuracy   & Accuracy\\
        & independently & combined with pool5\\
\hline
pool5   & \textbf{93.2\%}         & N.A                               \\
fc1000     & 87.8\%    & 89.0\%                               
\end{tabular}
\end{table}

\subsubsection{Choosing Hyperparameters:} 
\begin{figure*}
	\begin{center}
        \subfigure[]{\label{fig:8}\includegraphics[width=4.2cm,trim={1cm 0cm 3.3cm 0cm},clip]{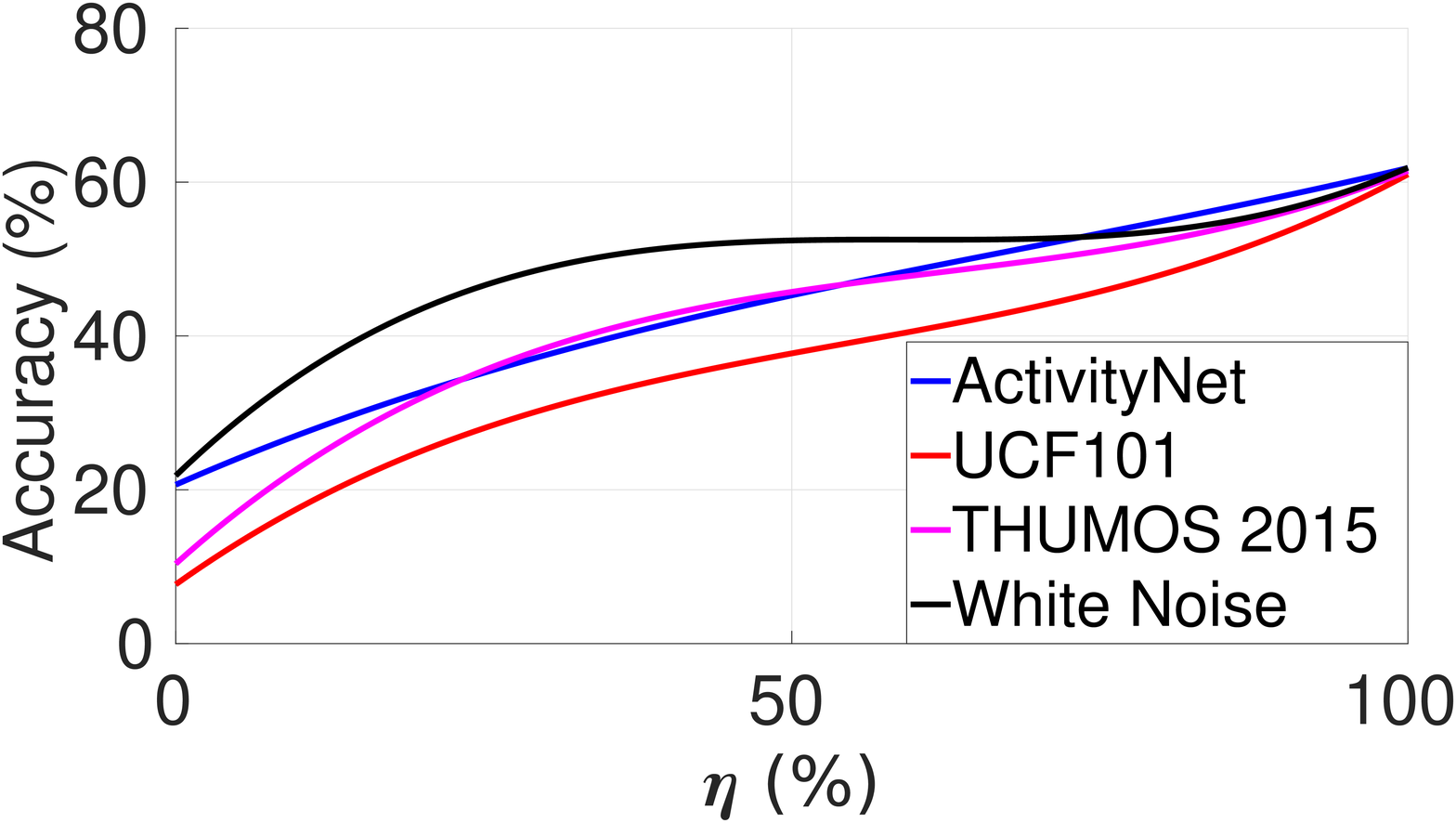}}
        \subfigure[]{\label{fig:3}\includegraphics[width=4.2cm,trim={1cm 0cm 4cm 0cm},clip]{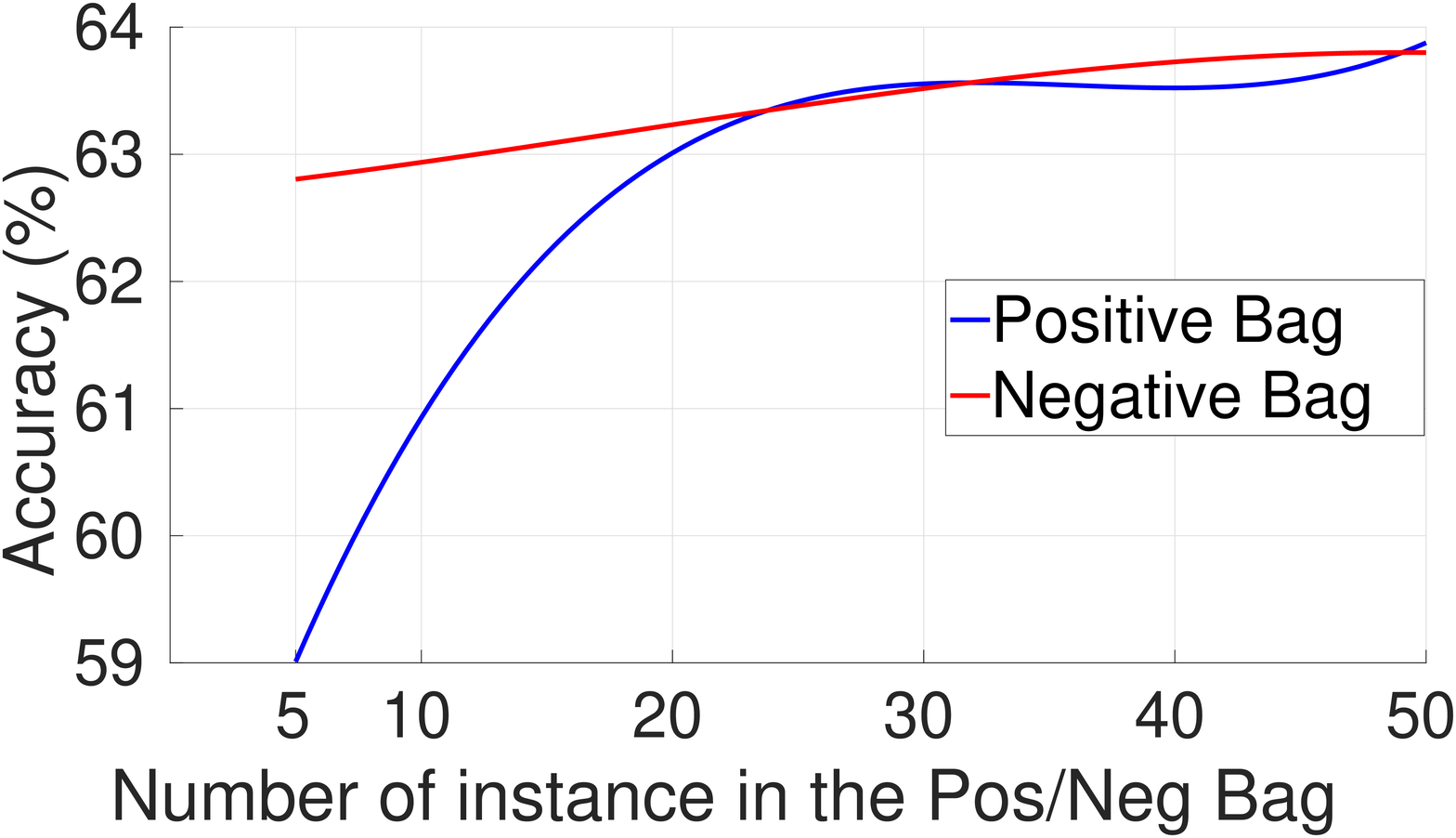}}
     \subfigure[]{\label{fig:4}\includegraphics[width=4.2cm,trim={1cm 0cm 4cm 0cm},clip]{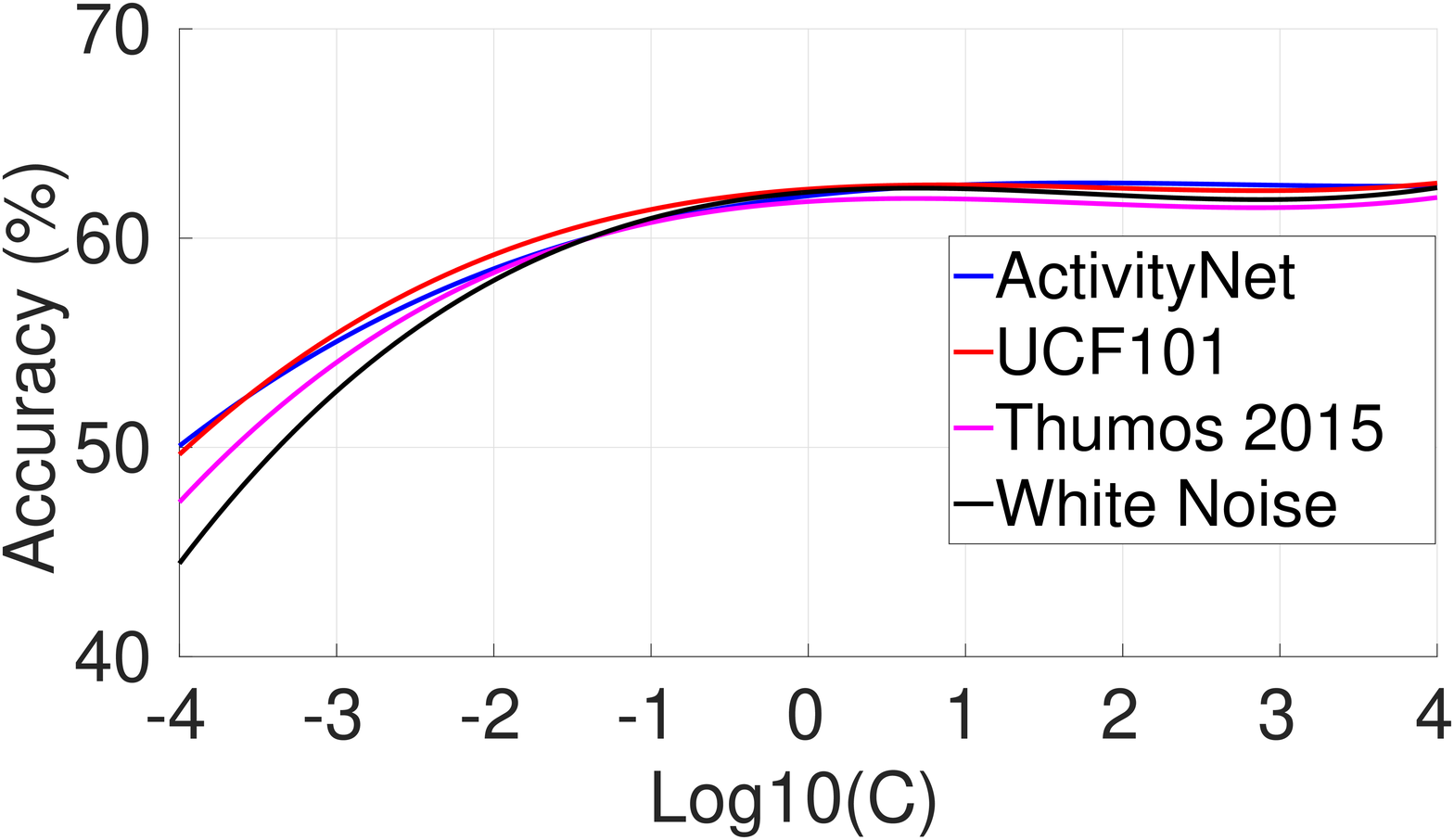}}
    \subfigure[]{\label{fig:5}\includegraphics[width=4.2cm,trim={1cm 0cm 4cm 0cm},clip]{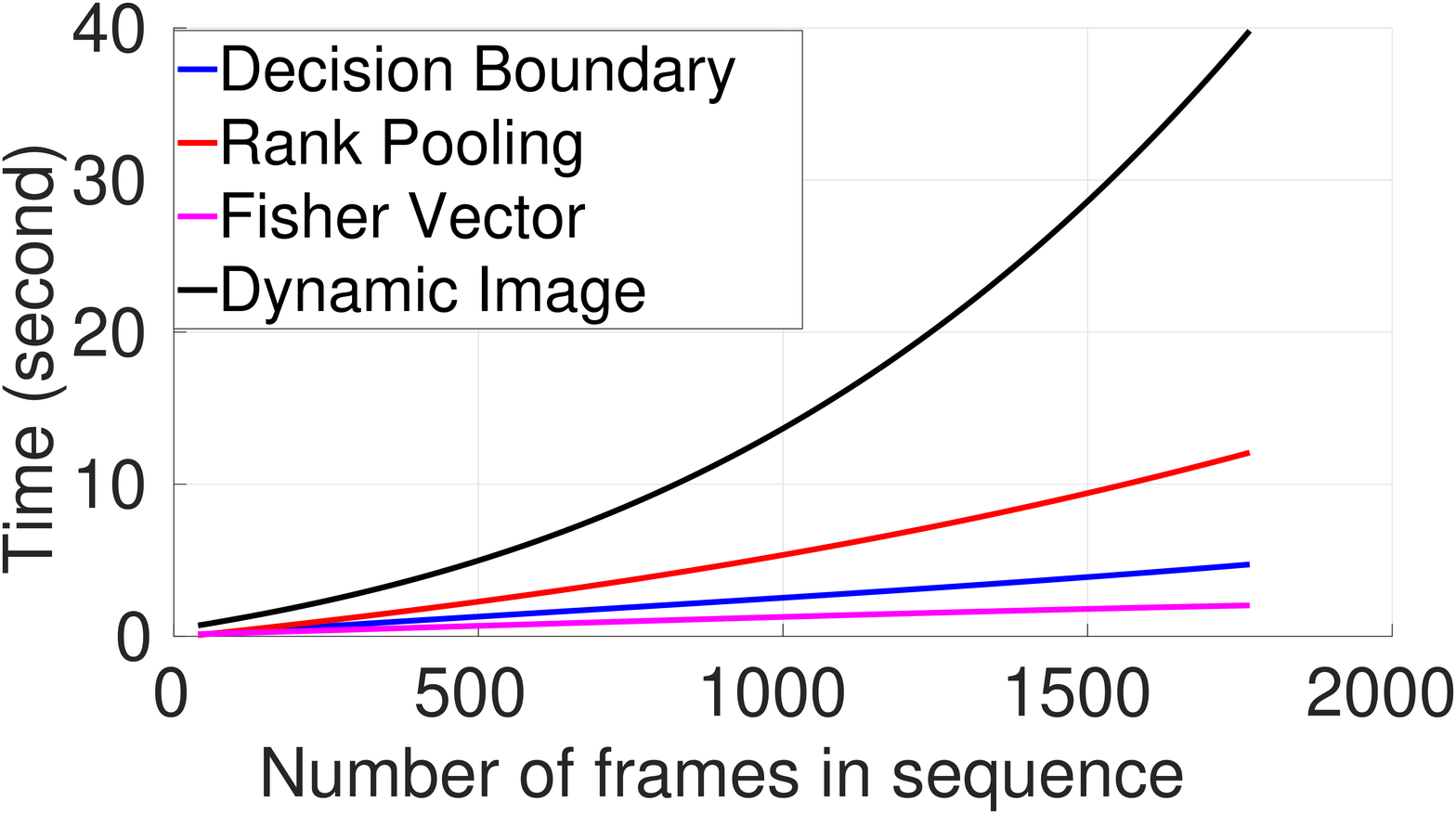}}        
	\end{center}
	\caption{Fig.~\ref{fig:8} shows an analysis of the influence of per-sequence accuracy threshold $\eta$ on the overall action recognition accuracy on HMDB51 split 1. Fig.\ref{fig:3} shows the accuracy of SVMP with different number of instance in Positive/Negative bag on HMDB51 split 1. Fig.~\ref{fig:4} shows the accuracy of SVMP descriptor against increasing slack regularization parameter $C_1$ used in our formulation~\eqref{eq:mil} on HMDB51 split 1. Fig.~\ref{fig:5} shows the running time of each popular algorithm.}
	\label{fig:all_plots}
\end{figure*}

The three important parameters in our scheme are (i) the $\eta$ parameter deciding the quality of an SVMP decision boundary, (ii) the respective $C_1=C$ used in Algorithm~\ref{alg1} when training for the decision boundary per sequence (to generate the SVMP descriptor) and (iii) the size of the positive and negative bags. To study the behavior of (i) and (ii), we plot in Figures~\ref{fig:8} and~\ref{fig:4}, classification accuracy when the per-sequence $C$ is increased from $10^{-4}$ to $10^{4}$ in steps and when $\eta$ is increased from 0-100\% and respectively. We repeat this experiment for all the different choices for negative bags. As is clear, increasing these parameters reduces the training error, but may lead to overfitting. However, Figure~\ref{fig:4} shows that increasing $C$ increases the accuracy of the SVMP descriptor, implying that the CNN features are already equipped with discriminative properties for action recognition. However, beyond $C=10$, a gradual decrease in performance is witnessed, suggesting overfitting. Thus, we use $C=10$ ( and $\eta=0.9$) in the experiments to follow. 

To decide the bag sizes for MIL, we plot in Figure~\ref{fig:3}, performance against increasing size of the positive bag, while keeping the negative bag size at 50 and vice versa; i.e., for the red line in Figure \ref{fig:3}, we fix the number of instances in the positive bag at 50; we see that the accuracy raises with the cardinality of the negative bag. A similar trend, albeit less prominent is seen when we repeat the experiment with the negative bag size, suggesting that about 30 frames per bag is sufficient to get a useful descriptor. For our MKL setup in~\eqref{eq:mkl}, we use $\beta_1=\beta_2=1$. 

\comment{
\begin{figure}[htbp]
	\centering
     \subfigure[]{\label{fig:3}\includegraphics[width=4cm,trim={2.5cm 0cm 4cm 1cm},clip]{figure/figure3.jpg}}
    \subfigure[]{\label{fig:5}\includegraphics[width=4cm,trim={2.5cm 0cm 4cm 1cm},clip]{figure/figure5.jpg}}
	\caption{Fig.~\ref{fig:3} shows the accuracy of SVMP descriptor against increasing slack regularization parameter $C_1$ used in our formulation~\eqref{eq:mil}. Fig.~\ref{fig:5} shows the running time of each popular algorithm.}
	\label{fig:eta}
\end{figure}
}

\subsubsection{SVMP vs NSVMP:} Recall that SVMP is a linear decision boundary, while NSVMP captures the non-linear boundary and thus could potentially be more representative of the action class. To check this claim, we evaluated their classification performance on HMDB and UCF101 datasets. The results are provided in Table~\ref{table:2}. Unlike the expectation, we find that the linear SVMP descriptors are significantly better than the NSVMP descriptors in terms of accuracy. This is perhaps due to the high-dimensionality of the  SVMP descriptors (fc6+pool5), while the NSVMP descriptors are lower dimensional (equal to the sum of the bag sizes). Although we tried classifying these descriptors via a non-linear kernel, their performance did not seem to improve, which suggests that perhaps the sequence RBF kernels used to generate these descriptors tend to easily overfit to the bags, thus making them less useful for sequence summarization. We further find that combining the SVMP and NSVMP kernels leads to some synergistic benefits as depicted in Table~\ref{table:2}. We observe this benefit both across datasets (HMDB and UCF101) and across CNN models for the same dataset (VGG-16 and ResNet-152 on UCF101).

\comment{
As introduce in the Section \ref{sec:proposed_method}, we apply both SVMP and NSVMP in our experiment and show the result in the table \ref{table:2}. In these two datasets, we find that the accuracy of NSVMP is lower than the one of SVMP. However, we found SVMP and NSVMP are complementary to each other, and thus, there is improvement after combination. In terms of the fusion strategy, we find that the homogeneous kernel linearization is quite useful, which could linearize the non-linear decision boundary and make it comparable with the linear decision boundary.}

\begin{table}[]
\centering
\caption{Accuracy comparison between SVMP and NSVMP on split-1 of the two datasets.}
\label{table:2}
\begin{tabular}{l|ll|l}
\hline
 \multicolumn{1}{l|}{} & \multicolumn{2}{c|}{VGG} & \multicolumn{1}{l}{ResNet152} \\ \hline
 & HMDB51 & UCF101 & UCF101\\\hline
SVMP     & 63.79\%& 91.58\%  & 92.20\%     \\
NSVMP & 53.41\%& 79.11\%   & 80.30\% \\
Combination   & \textbf{64.51\%}&  \textbf{92.20\%} &\textbf{93.50\%}    
\end{tabular}
\end{table}

\subsubsection{Comparison to Standard Pooling Schemes} 
In Table~\ref{table:3}, we compare SVM pooling against average pooling and fusion by a linear SVM as suggested in~\cite{simonyan2014two,feichtenhofer2016convolutional}. We see that the combination of SVMP and NSVMP leads to significant improvements on both the datasets. Specifically, it leads to about 5\% improvement on the HMDB dataset. In \cite{feichtenhofer2016convolutional}, they apply average pooling on the top of the CNN features, compared with which, using SVMP brings 6.3\% and 1.5\% improvement in HMDB51 and UCF101 respectively. 

\begin{table}[]
\centering
\caption{Accuracy comparison between SVM Pooling and standard pooling methods on split1 of the two datasets.}
\label{table:3}
\begin{tabular}{lll}
\hline
                  & HMDB51  &UCF101 \\\hline
Spatial Stream    & 47.06\% &83.40\% \\
Temporal Stream   & 55.23\% &87.20\% \\
Two-stream 		  & 58.17\% &91.80\%\\
Two-stream with SVM &59.80\%&91.50\%\\
SVMP + NSVMP & \textbf{64.51\%}&\textbf{93.50\%}     
\end{tabular}
\end{table}

\subsubsection{Complementarity to Hand-crafted Features} 
Combining CNN-based features with hand-crafted dense trajectories-based features~\cite{wang2013dense} (such as using IDT-FV) is often found to improve the performance of state-of-the-art methods. In Table \ref{table:4}, we show the accuracy before and after combining SVMP and NSVMP descriptors with IDT-FV features. The result shows an improvement of about  6\% and 1\% respectively on the HMDB-51 and UCF101 datasets, which we believe is significant.

\begin{table}[]
\centering
\caption{Accuracy comparison before and after including dense trajectory features on split1 in two datasets.}
\label{table:4}
\begin{tabular}{llll}
\hline
                           & HMDB51    & UCF101 \\\hline
IDT-FV                     & 60.10\%   & 85.32\% \\
SVMP + NSVMP          & 64.51\%   & 93.50\%       \\
SVMP + NSVMP + IDT-FV & \textbf{70.60\%}   &\textbf{94.60\%}       
\end{tabular}
\end{table}

\subsubsection{Comparisons to the State of the Art} 
In Table \ref{table:5}, we compare our best result (from the last experiment) against the state-of-the-art results on HMDB51 and UCF101 datasets. Recall that, on HMDB-51 dataset, we use the VGG-16 model in the two-stream framework and apply SVM pooling on top of CNN features from 'pool5' and 'fc6', combined with the improved dense trajectory features. As is clear from the table, we outperform the recent state of the art on HMDB-51 by about 0.3\%. On the UCF101 dataset, we use pool5 features from a ResNet-152 model to generate our descriptors and we achieve the same state-of-the-art performance as the very recent~\cite{feichtenhofer2016spatiotemporal}. Given the simplicity and generality of our pooling scheme, we believe these results are significant considering the difficulties of recognition in these datasets as exemplified by the recent trend in improvements produced by the state-of-the-art methods. We also note that our pooling scheme is general and independent of any specific CNN architecture. In contrast, ~\cite{feichtenhofer2016spatiotemporal} uses a sophisticated spatio-temporal residual network with interconnections between the two-streams which can be difficult to train. 


\begin{table}[]
\centering
\small
\caption{Comparison to the state of the art. Results on HMDB51 and UCF101 using mean classification accuracy of the best performance over 3 splits. $^*$Note that as we use an additional small negative bag, our results are not exactly comparable to some of the-state-of-the-art methods. This additional bag does not favor any positive class and it is available to all other methods.}
\label{table:5}
\begin{tabular}{lll}
\hline
Method & HMDB51 & UCF101  \\\hline
Two-stream \cite{simonyan2014two}                       & 59.4\%  & 88.0\% \\
Very Deep Two-stream Fusion \cite{feichtenhofer2016convolutional} & 69.2\% & 93.5\%  \\
Temporal segment networks\cite{Wang2016}                & 69.4\% & 94.2\% \\
Composite LSTM \cite{srivastava2015unsupervised}        & 44.0\% & 84.3\% \\
IDT+FV \cite{wang2013action}                            & 57.2\% & 85.9\%\\
IDT+HFV \cite{peng2016bag}                              & 61.1\% & 87.9\% \\
TDD+IDT \cite{wang2015action}                           & 65.9\% & 91.5\% \\
DT+MVSV \cite{cai2014multi}                             & 55.9\% & 83.5\% \\
Dynamic Image + IDT-FV \cite{bilen2016dynamic}          & 65.2\% & 89.1\% \\
Dynamic Flow + IDT-FV\cite{dynamic_flow}                            & 67.4\% & 91.3\%\\
Generalized Rank Pooling~\cite{grp}                     & 67.0\% & 92.3\%\\
Spatial-temporal ResNet \cite{feichtenhofer2016spatiotemporal} &70.3\% & \textbf{94.6}\% \\\hline							
Ours (SVMP + NSVMP + IDT-FV)$^*$                       &\textbf{70.6\%} &\textbf{94.6\%}\\\hline
\end{tabular}
\end{table}

\subsection{Running Time Analysis}
As we solve an SVM objective, our algorithm is expected to be more expensive than a simple average pooling scheme typically used~\cite{simonyan2014two}. In Figure~\ref{fig:5}, we compare the time it took on average to generate SVMP descriptors for an increasing number of frames in a sequence on the UCF101 dataset. To compare our performance, we also plot the running times for some of the recent and popular pooling schemes such as~\cite{bilen2016dynamic,fernando2015modeling} and the Fisher vector scheme~\cite{wang2013action}. The plot shows that while our scheme is slightly more expensive than standard Fisher vectors (using the VLFeat\footnote{http://www.vlfeat.org/}, it is significantly cheaper to generate SVMP descriptors in contrast to some of the recent popular pooling methods.

%% file: conclude.tex
\section{Conclusion}
\label{sec:conclude}
In this paper, we presented a simple, efficient, and powerful pooling scheme, SVM pooling, for summarizing actions in videos. We cast the pooling problem in a multiple instance learning framework, and seek to learn useful decision boundaries on the frame level features from each sequence against a randomly chosen set of unrelated action sequences. We provide an efficient scheme that jointly learns these decision boundaries and the action classifiers on them. We also extended the framework to deal with non-linear decision boundaries. Extensive experiments were showcased on two challenging benchmark datasets, HMDB and UCF101, demonstrating state-of-the-art performance.
\comment{
It summarize actions in video sequences by filtering useful features from a bag of per-frame CNN features. And We propose to use the output of SVM pooling as a descriptor for representing the sequence, namely the SVM Pooled (SVMP) descriptor. Considering SVMP descriptor as a better representation than CNN features, we extended SVMP to NSVMP to deal with the  non-linear sequences. Extensive experiments were implemented on two popular benchmark datasets: HMDB51 and UCF101, which clearly show the advantage and effectiveness of SVMP compared with the traditional pooling scheme on CNN features. More importantly, after combining with the hand-crafted local features, we reach the state-of-the-art result on these two datasets.
}